\documentclass[letterpaper, 10 pt, conference]{ieeeconf}

\IEEEoverridecommandlockouts
\usepackage[T1]{fontenc}
\usepackage{amsmath}
\usepackage{amssymb}
\usepackage{booktabs}
\usepackage{graphicx}
\usepackage{xcolor}
\usepackage{url}
\usepackage{pifont}
\usepackage{capt-of}
\usepackage{algorithm}
\usepackage[noend]{algpseudocode}
\usepackage{newtxtext}
\usepackage{listings}
\usepackage{tcolorbox}
\tcbuselibrary{listings}

\newcommand{\methodshort}{PARTS}
\newcommand{\methodlong}{Policy Adaptation with RL on Targeted Subtasks}

\definecolor{AbstractIntroRevision}{RGB}{0,76,166}

\title{\LARGE \bf
From Pretraining to Proficiency: Real-World Subtask RL for Long-Horizon Manipulation with Minimal Human Intervention
}

\author{%
  Sichang Su$^{1}$, Benjamin Yang$^{2}$, Zhiyun Deng$^{1}$, Boyuan Liang$^{3}$,\\
  Yip Fun Yeung$^{2}$, Zelin Wang$^{2}$, and Lingfeng Sun$^{2}$%
  \thanks{$^{1}$UT Austin. $^{2}$Autel US. $^{3}$UC Berkeley.}%
  \thanks{\raggedright Corresponding authors: Sichang Su (\mbox{sichang\_su@utexas.edu})
    and Lingfeng Sun (\mbox{lingfengsun1996@gmail.com}).}%
    \\[0.5em]
  {\textcolor{blue}{\url{https://destiny000621.github.io/PARTS/}}}%
}

\IEEEaftertitletext{%
  \vspace{-0.6\baselineskip}%
  \begin{center}%
  \includegraphics[width=\textwidth]{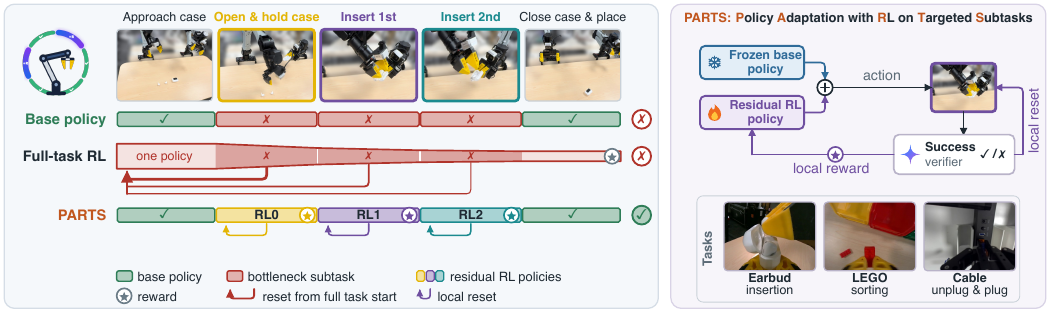}%
  \captionof{figure}{\methodshort{} overview. A pretrained policy completes most subtasks of a long-horizon task but fails at a few bottlenecks; full-task success requires every subtask to succeed. Full-task RL fine-tunes a single policy using a sparse reward delivered only upon full-task success. Each bottleneck failure triggers an episode restart, so progressively fewer rollouts reach later subtasks (narrowing bar).
  In contrast, \methodshort{} retains the frozen base policy for subtasks it already handles reliably and trains one residual policy per bottleneck, with a local reward for every attempt.}%
  \label{fig:overview}%
  \end{center}%
  \vspace{2pt}%
}

\begin{document}

\maketitle
\thispagestyle{empty}
\pagestyle{empty}

\begin{abstract}
A pretrained robot foundation policy may execute most of a long-horizon task yet repeatedly fail at a few critical subtasks. Collecting additional full-task demonstrations for supervised fine-tuning (SFT) requires operators to repeat behaviors the policy already performs well. Reinforcement learning (RL) fine-tuning offers a promising path to bridge this gap, but existing approaches struggle to solve long-horizon tasks using only sparse rewards. We present \methodshort{} (Policy Adaptation with RL on Targeted Subtasks), a real-world subtask RL framework that concentrates practice at these bottlenecks while allowing training rollouts to proceed with minimal human intervention. The frozen pretrained policy supplies nominal actions throughout execution, while agent-generated selectors and success verifiers activate residual corrections and provide local outcome rewards. These rewards support learning from successful subtasks even when complete-task successes are scarce. Training combines online RL with success-reweighted retraining, and each retrained residual policy is redeployed to collect further experience. Humans identify bottlenecks during setup and perform physical resets when needed. On bimanual YAM and single-arm Franka tasks, \methodshort{} improves complete-task success from 32\% to 61\% and from 50\% to 95\%, respectively, using tens of minutes of real-world RL rollouts per task on average.
Compared with existing real-world RL fine-tuning methods, \methodshort{} raises full-task success by more than 25\% under the same robot-rollout budget while requiring less human involvement.

\end{abstract}

\section{INTRODUCTION}

Pretrained robot policies offer useful behaviors for adapting to new manipulation tasks. Recent vision-language-action (VLA) models and world-action models (WAMs) draw on large robot datasets, visual and semantic knowledge, and video prediction to produce increasingly capable policies~\cite{pmlr-v305-black25a,physicalintelligence2026pi07,ye2026dreamzero,li2026novaflow}. A target task can nevertheless demand changes in grasp strategy, contact behavior, spatial arrangement, or coordination between successive actions. Supervised fine-tuning (SFT) on additional demonstrations can address these differences. For a long-horizon task, collecting complete demonstrations repeatedly incurs human effort even for behaviors that already work. We study how to make better use of a pretrained policy's existing capabilities while learning the changes needed for reliable target-task execution.

Our starting observation is that failures in the tasks we study concentrate at a few consequential subtasks. These bottlenecks include high-precision operations, such as earbud insertion, and preparatory subtasks whose terminal states affect subsequent execution. For example, a robot may successfully pick up an earbud but hold it in a pose that makes insertion difficult. Adaptation may therefore target both precision-critical motions and earlier actions that establish suitable grasps or placements, while retaining the reliable behaviors of the pretrained policy.

Real-world RL provides a way to improve existing behaviors through interaction, including by learning residual corrections to a fixed policy~\cite{ankile2025residual}. If learning uses only full-task success, an improved intermediate behavior can still receive no positive reward when a later step fails. Repeated early failures also reduce opportunities to practice later subtasks. Local episodes with verifiable outcomes can provide successful experience before complete-task execution becomes reliable, provided the initial policy supports productive local exploration. Prior work uses planned subgoals for online adaptation~\cite{fang2022ptp} and refines selected task phases~\cite{xu2026rl,zheng2026torl}. Our focus is on organizing such local learning to adapt a pretrained policy across the bottlenecks of a long-horizon physical task. \textit{We train on bottleneck subtasks and measure progress by success of the complete task.}

Local practice must integrate with full-task execution. Each subtask requires reachable entry states and a success criterion that captures readiness for the next stage. The system must manage continuation, retries from the current state, and physical reset requests, while selecting bottlenecks for improvement and deciding when to deploy updated policies.
The framework must therefore coordinate episode supervision, task execution, and policy improvement with minimal human involvement.

We present \methodshort{} (\methodlong{}), a real-world subtask RL framework for adapting pretrained robot policies, as shown in Fig.~\ref{fig:overview}. A frozen base policy supplies nominal actions throughout execution, and lightweight, bounded residual policies modify those actions within selected bottlenecks. Given human-identified bottlenecks, coding agents construct executable policy selectors and success verifiers; VLMs act as language-conditioned visual perception modules. The selectors govern residual activation, and the verifiers supply sparse local outcome rewards. Online RL improves the residuals with newly collected robot experience. Success-reweighted retraining emphasizes successful local experience in the accumulated data, and the retrained policies are deployed to collect further rollouts. Retraining thus affects both the policy and the experience available for further learning. Training rollouts require no human action corrections, switching decisions, or outcome labels. Humans specify bottlenecks, assist with setup, and perform physical resets when needed.
We instantiate our framework using $\pi_{0.5}$ as the backbone foundation model, a state-of-the-art generalist policy that has demonstrated strong performance across diverse manipulation tasks~\cite{pmlr-v305-black25a}.

We evaluate \methodshort{} on two bimanual long-horizon tasks with YAM and one single-arm long-horizon task with Franka; earbud and cable insertion require millimeter-level precision, while LEGO sorting involves bricks that are out of distribution for the base policy. Full-task success increases from 32\% to 61\% on YAM and from 50\% to 95\% on Franka, requiring only tens of minutes of real-world RL rollouts per task on average.
Compared with existing real-world RL fine-tuning methods, \methodshort{} raises full-task success by more than 25\% while requiring less human involvement.

Our contributions are:
\begin{itemize}
    \item A formulation for adapting pretrained robot policies through RL at selected bottlenecks, using local outcome supervision while evaluating reliability over complete long-horizon execution.
    \item \methodshort{}, a framework connecting executable subtask supervision, residual RL, and success-reweighted retraining with redeployment, enabling training rollouts without human corrections, switching decisions, or reward labels.
    \item Experiments on the bimanual YAM and single-arm Franka platforms showing that \methodshort{} boosts full-task success even from weak initial policies under a limited real-world rollout budget.
\end{itemize}

\section{RELATED WORK}
\begin{figure*}[t]
  \centering
  \includegraphics[width=\textwidth]{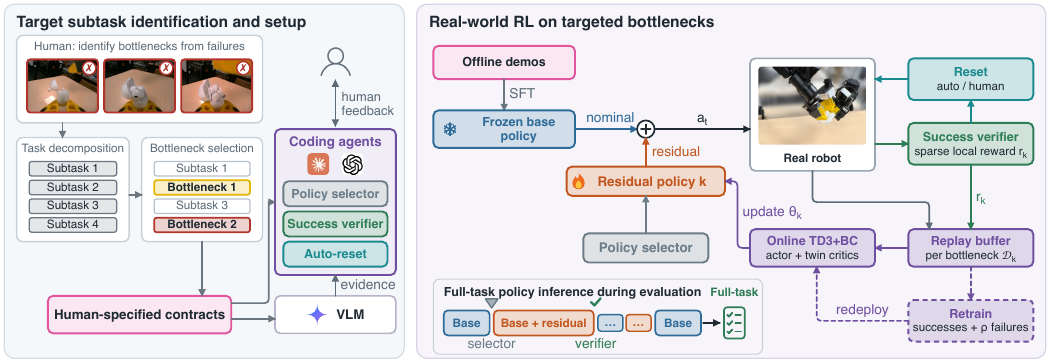}
  \caption{\methodshort{} architecture. \emph{Left:} Humans identify bottlenecks from base-policy failures and specify contracts. Coding agents implement selection, verification, and reset programs.
  \emph{Right:} Residual policies correct the frozen base policy. Local rewards and per-bottleneck replay support online TD3+BC and retraining on all successes plus a fraction $\rho$ of failures.
  Retrained policies resume online learning. At evaluation, \methodshort{} enables the switch between the base and residual policies to complete the full task.}
  \label{fig:architecture}
  \vspace{-9pt}
\end{figure*}

\begin{table}[t]
  \centering
  \caption{Human involvement during RL data collection, as published for the
  baselines and \methodshort{}. \ding{51}: autonomous; \ding{55}: requires a
  human. \emph{Rollout}: corrective interventions or handoff decisions during
  RL rollouts; \emph{Reward}: success labeling; \emph{Reset}: scene
  restoration.}
  \label{tab:human_involvement}
  \small
  \setlength{\tabcolsep}{6pt}
  \begin{tabular}{lccc}
    \toprule
    Method & Rollout & Reward & Reset \\
    \midrule
    DSRL~\cite{wagenmaker2025steering} & \ding{51} & \ding{55} & \ding{55} \\
    EXPO-FT~\cite{dong2026expo} & \ding{55} & \ding{51}$^{\dagger}$ & \ding{55} \\
    RLT~\cite{xu2026rl} & \ding{55}$^{\ast}$ & \ding{55} & \ding{55} \\
    \midrule
    \methodshort{} (ours) & \ding{51} & \ding{51}$^{\ddagger}$ & \ding{51}$^{\ddagger}$ \\
    \bottomrule
  \end{tabular}
  \par\smallskip
  \parbox{0.97\columnwidth}{\footnotesize
  $^{\ast}$A human hands control from the VLA to the RL policy in every
  episode; corrective interventions are optional.
  $^{\dagger}$Rule-based detectors, with a human override key in the released
  code.
  $^{\ddagger}$Automatic on LEGO and cable; on earbuds a human checks the
  verifier's labels, and a human resets when the robot cannot restore the
  scene.}
\end{table}

\subsection{Real-World RL in Long-Horizon Tasks}
Unlike sim-to-real RL pipelines~\cite{shi2026beyond,zang2025rlinfvla}, which train with RL in simulation before hardware deployment, real-world RL updates a policy through physical interaction, where data collection, resets, and unproductive exploration are costly. Prior systems improve efficiency through demonstrations~\cite{lei2026performant,luo2024serl}, residual control~\cite{johannink2019residual,ankile2025residual}, learned reward functions~\cite{biza2025goalcontrastive}, and human intervention~\cite{luo2025precise}.
Long-horizon tasks introduce an additional challenge because rewards are often sparse, prior work decomposes them
into shorter-horizon goals or reusable skills, composing demonstrated
primitives~\cite{fang2022ptp}, discovering skills from prior
experience~\cite{zhu2022bottom,shi2022skill}, or sharing parameters across
related tasks~\cite{sun2022paco}.
More recently, pretrained VLA policies have become increasingly capable of executing complex tasks, shifting the problem from learning behaviors from scratch toward improving already capable policies~\cite{kim2024openvla,pmlr-v305-black25a,physicalintelligence2026pi07}.
Online RL has consequently been used to further adapt pretrained VLA policies to real-world tasks~\cite{guo2025improving,xu2026rl,dong2026expo}.
However, these methods emphasize global policy improvement. PARTS instead adopts \emph{failure-localized policy repair}, concentrating real-world RL practice on the few bottleneck subtasks that limit long-horizon success while retaining already reliable behaviors.
Several recent systems also restrict learning to part of a task~\cite{xu2026mori, chen2026bora, hausdorfer2026data}. \methodshort{} likewise targets base-policy failures, but delimits bottlenecks through executable contracts rather than action variance or a fixed contact phase. Automatic verifiers provide local outcome rewards instead of intervention-derived rewards.

\subsection{Human Supervision for Real-World Policy Adaptation}
Human supervision is commonly used in real-world policy learning to guide exploration and prevent costly failures through corrective actions~\cite{celemin2019reinforcement}, policy takeover or switching \cite{luo2025precise, xu2026rl}, and success/failure labeling \cite{pamies2023autonomous,xu2026rl}.
Although effective, these approaches rely on human monitoring or intervention during task execution, limiting their scalability.
Recent systems automate parts of this supervision:
UniIntervene~\cite{deng2026uniintervene} detects value degradation and
retrieves recovery behaviors, and Robot Trains Robot~\cite{hu2025robot}
automates protection, failure detection, scheduling, and resets, in both
cases for a policy that practices the whole task. \methodshort{} instead
confines practice to bottleneck subtasks, so that outcome labeling and
resets become local problems handled by executable verifiers and reset
programs, and RL rollouts need no human corrections or switching decisions.

\section{METHOD}
\label{sec:method}

\subsection{Problem Statement}
\label{sec:problem}

We consider fine-tuning a pretrained VLA policy $\pi_{\mathrm{VLA}}$
($\pi_{0.5}$ in this work) with real-world RL. $\pi_{\mathrm{VLA}}$ is a
generalist trained on a diverse demonstration corpus and conditioned on a language
instruction~\cite{pmlr-v305-black25a}. Like most modern VLAs, it uses action
chunking: at each replan it predicts a chunk
$\bar A_t=(\bar a_t,\ldots,\bar a_{t+C-1})$ of $C$ future actions and executes
a prefix of $E\leq C$ actions before predicting again~\cite{pmlr-v305-black25a}. Observations $o_t$ consist of
multi-view RGB images and the proprioceptive state $p_t$. For tasks on which
$\pi_{\mathrm{VLA}}$ exhibits limited zero-shot performance, we assume access to a small offline dataset of
expert demonstrations, $\mathcal{D}_{\text{exp}}$, obtained through human
teleoperation and from open-source datasets. SFT on
$\mathcal{D}_{\text{exp}}$ yields the task-specific base policy, denoted
$\pi_0$, from which all RL methods in this paper start.

The target tasks are long-horizon. We model an episode as a sequence of $N$
subtasks $\sigma_1,\ldots,\sigma_N$, for example opening a case, inserting a
first earbud, and inserting a second. Subtask $i$ is entered from a set of
states $\mathcal E_i\subset\mathcal S$ and ends when its outcome criterion
$\phi_i:\mathcal S\to\{0,1\}$ is evaluated at exit or a time limit expires. The
only reward is a sparse binary signal $R\in\{0,1\}$ for task completion,
provided by a success verifier or a human at the end of the episode; no dense
shaping is available. Because the task succeeds only if every subtask
succeeds, this terminal reward factorizes as
\begin{equation}
 R=\prod_{i=1}^{N}\phi_i ,
 \label{eq:task_reward}
\end{equation}
although a learner that observes only $R$ never sees the individual
$\phi_i$. If the base policy
completes subtask $i$ with probability $p_i$ from the states it typically
enters, then $\Pr[R=1]\approx\prod_i p_i$, so a few subtasks with low $p_i$
bound complete-task success even when the remaining stages are reliable. We
call these \emph{bottleneck subtasks} and write $\mathcal K\subseteq\{1,\ldots,N\}$
for their index set, $K=|\mathcal K|$. A bottleneck may begin before the visible failure. When the terminal state of $\sigma_{i-1}$ determines the difficulty of $\sigma_i$, the bottleneck extends into $\sigma_{i-1}$, whose success criterion requires a configuration suitable for executing $\sigma_i$.
For example, case preparation succeeds only when the lid is open and the case is held in an orientation that facilitates earbud insertion. Bottlenecks may form
an ordered chain or a set of alternatives, and the same bottleneck may recur
within one episode, such as a grasp that repeats for every object. The entry
distribution $\mu_i$ of subtask $i$ is induced by executing
$\sigma_1,\ldots,\sigma_{i-1}$ from the task's initial conditions; a subtask
may also be entered from a restaged distribution $\hat\mu_i$ prepared by a
reset procedure, which need not equal $\mu_i$. The objective is to maximize the full-task success rate.

This setting is hard for two reasons. First, the reward~\eqref{eq:task_reward}
is sparse over a long horizon, and when the base policy rarely completes the
task, most rollouts return no signal at all. Second, real-robot training time
is limited, so a budget spent uniformly over the task leaves few attempts at
the bottlenecks.

\subsection{\methodlong{}}
\label{sec:parts_core}

Fig.~\ref{fig:architecture} summarizes our recipe for adapting a pretrained policy to a long-horizon task with real-world RL. The core idea is to spend robot
interaction only where the pretrained policy fails. Fine-tuning the whole task
online would revisit subtasks the base policy already performs and would
receive the sparse terminal reward~\eqref{eq:task_reward} only when every stage
succeeds. Instead, we keep the VLA frozen: it executes every subtask, supplies
reference action chunks and visual features, and defines the neighborhood in
which small residual policies may act. We first fine-tune the VLA on task
demonstrations and identify the bottlenecks where it still fails. For each
bottleneck we then train a lightweight residual actor-critic with
online TD3+BC, bounded to a few action dimensions around the VLA's reference chunk and
rewarded by that subtask's own outcome, so that each attempt yields a usable
signal. Executable programs authored by coding agents determine when to activate a residual policy,
whether an attempt has succeeded, and how to reset, enabling training rollouts without human corrections or manual policy switching. Periodic
success-reweighted retraining consolidates the rare successes and redeploys
the residual to collect further experience. At inference, the fixed residuals
are activated at their entries and hand control back to the base policy at
their exits. This design turns real-world RL into targeted refinement of a few
behaviors while the rest of the task retains the reliability of the pretrained
model.

\paragraph{Residual RL}

Each time the frozen base policy is queried, it supplies the nominal chunk $\bar A_t$. For
bottleneck $k\in\mathcal K$, a residual actor $f_{\theta_k}(z_t,p_t,\bar A_t)$
outputs a normalized residual chunk $U_{k,t}\in[-1,1]^{C\times d_k}$,
conditioned on visual features $z_t$ extracted by the base policy,
proprioception $p_t$, and the nominal chunk. A binary coordinate mask $M_k$ and physical bounds $B_k$, specified per
bottleneck (Sec.~\ref{sec:scaffolding}), restrict the correction.
Writing $g_t\in\{0\}\cup\mathcal K$ for the active bottleneck, with zero
denoting nominal execution, the command is
\begin{equation}
 a_t=\mathcal C\!\left(\bar a_t+
 \sum_{k\in\mathcal K}\mathbf 1[g_t=k]\,
 B_k\odot M_k\odot u_{k,t}\right).
 \label{eq:residual}
\end{equation}
Here $u_{k,t}$ is the current row of $U_{k,t}$ after temporal smoothing,
embedded in the robot's action coordinates, and $\mathcal C$ applies the
command constraints. The mask confines RL to the action dimensions relevant to a
bottleneck, such as gripper openness or end-effector translation, while every
other dimension follows the base policy. Untrained actors output a zero residual,
and exploration adds Gaussian noise to the normalized residual before clipping.
The residual learning formulation~\cite{ankile2025residual,xu2026rl} requires only nominal action chunks and an observation
representation, so it is agnostic to the base policy's architecture.

\paragraph{TD3+BC learner}

Each bottleneck owns a residual actor, twin critics, and a replay buffer
$\mathcal D_k$ whose transitions store the RL state $s=(z,p)$, the nominal
chunk $\bar A$, the executed residual chunk $U$, the chunk's per-step rewards
$r_\tau$, whose sum over an attempt equals the local reward $r_k$, and the
terminal flag $d$; we drop the bottleneck index below. We train with chunk-level
TD3~\cite{fujimoto2018addressing} and behavior-cloning
regularization~\cite{fujimoto2021minimalist}. The critics $Q_j(s,U)$ estimate the
value of a residual chunk in a state and are trained by temporal-difference
learning over the $C$-step chunk. With bars denoting target networks, Gaussian
target noise $\epsilon$, the coordinate mask $M$, and a minibatch
$\mathcal B\subset\mathcal D_k$,
\begin{align}
 U' &= M\odot\operatorname{clip}
       \bigl(f_{\bar\theta}(s',\bar A')+\epsilon,-1,1\bigr),\nonumber\\
 y &= \sum_{\tau=0}^{C-1}\gamma^{\tau} r_\tau
       +(1-d)\gamma^C\min_{j=1,2}Q_{\bar\psi_j}(s',U'),\label{eq:target}\\
 \mathcal L_Q &= \sum_{j=1}^{2}
       \mathbb E_{\mathcal B}\bigl[(Q_{\psi_j}(s,U)-y)^2\bigr].\nonumber
\end{align}
The actor maximizes the critic's value while staying close to corrections that
worked: it clones the executed residuals of successful episodes
$\mathcal B^{+}$ and, optionally, penalizes nonzero residuals of failed episodes
$\mathcal B^{-}$, which anchors failed attempts back to the nominal policy. With
$\widehat U$ denoting the actor's masked prediction,
\begin{equation}
\begin{split}
 \mathcal L_\pi={}&-\lambda_Q\mathbb E_{\mathcal B}
                  [Q_{\psi_1}(s,\widehat U)]\\
 &+\lambda_+\langle\|\widehat U-U\|^2\rangle_{\mathcal B^+}
  +\lambda_-\langle\|\widehat U\|^2\rangle_{\mathcal B^-},
\end{split}
\label{eq:actor}
\end{equation}
where the weights $\lambda$ balance value improvement, imitation, and
anchoring. Because conditioning on $\bar A$ and regularizing toward executed
residuals can let the actor copy rather than improve, we apply reference
dropout, zeroing $\bar A$ for a random subset of each batch~\cite{xu2026rl}.

\begin{figure*}[t]
  \centering
  \includegraphics[width=1\textwidth]{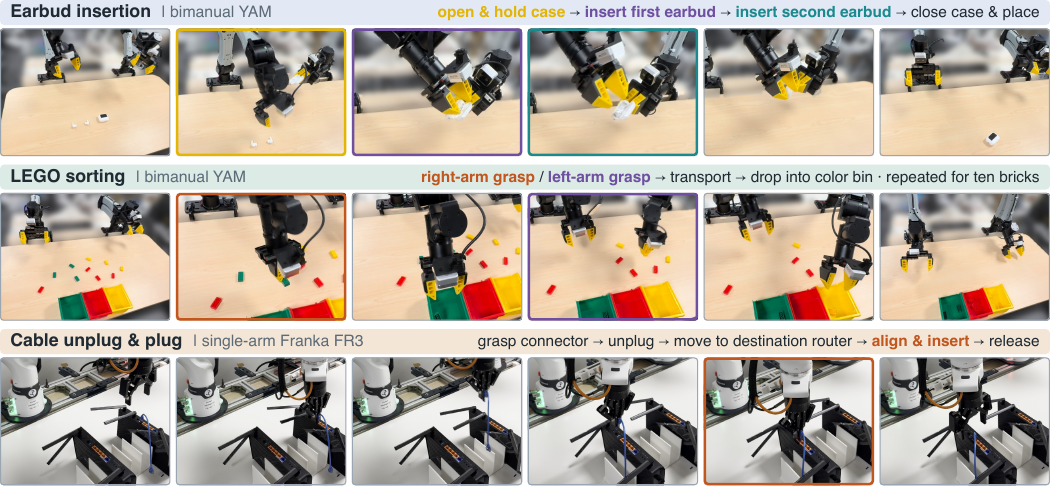}
  \caption{Tasks on bimanual YAM and single-arm Franka FR3. Colored borders mark states produced by a
  targeted residual policy, in the color of its subtask name in the row header.
  \emph{Top:} earbud insertion, where inserting earbuds requires high precision.
  \emph{Middle:} LEGO sorting, where the grasp of each small brick is the
  recurring bottleneck because the brick size is out-of-distribution for the base policy. \emph{Bottom:} cable insertion requires precise manipulation. }
  \label{fig:tasks}
  \vspace{-9pt}%
\end{figure*}

\subsection{Agentic Scaffolding}
\label{sec:scaffolding}

A human identifies the bottlenecks $\mathcal K$ from real-robot evaluations of
$\pi_0$ and writes a \emph{contract} for each: its entry set $\mathcal E_k$,
the correctable action coordinates and their bounds, its outcome criterion
$\phi_k$, and a motion budget, with a handful of subtask demonstrations.
Each bottleneck is then supervised by its own local reward $r_k=\phi_k$; the
complete-task outcome $R$ serves only for evaluation. Turning a contract into
repeatable practice requires attempts that start at reachable states, end
with a trustworthy label, and can be repeated. \methodshort{} implements these
functions as three executable programs (a policy selector, a success
verifier, and a reset policy) that run alongside the control loop. Coding agents (Claude Fable 5 and GPT-6 Astra) author these programs using the contracts, subtask demonstrations, robot interfaces, and recorded rollouts~\cite{liang2023code,elmaaroufi2026rho,fu2026cap}.
Humans review the programs' decisions on labeled episodes, and the agents revise the code accordingly. Once deployed, the programs run as ordinary code and do not generate motor commands. Table~\ref{tab:human_involvement} summarizes the human involvement that remains during RL data collection, compared with the baselines.
Their perceptual
evidence comes from promptable segmentation with
SAM3~\cite{carion2026sam}, which supplies object masks and
locations for geometric predicates, and from a VLM (Gemini-3.7-flash) that
answers asynchronous queries about specified object states, as language models have been used to resolve partially observable task state~\cite{sun2024interactive}.

\textbf{Policy Selector.}
The selector outputs $g_t$ from images, proprioception, and motion predicates
under the task structure of the contracts, either an ordered sequence of
residuals or a choice among those whose entry conditions hold, in both cases
allowing repeated activation when an entry recurs. It activates residual $k$
only when the observations support that bottleneck's entry conditions and
otherwise leaves the base policy in control.

\textbf{Success Verifier.}
The verifier produces the local reward. Once human-defined event gates are
satisfied, such as the gripper opening after an insertion, it tests the
contract's postcondition, requiring persistence across observations when a
stable outcome matters, and returns one for success and zero for failure. An uncertain verdict requests a human label, and a human can override
an automatic label. During evaluation, verified subtask success triggers a handoff from the residual policy to the base policy, allowing full-task execution to continue and subsequent residual policies to be activated as needed.

\textbf{Auto-Reset Policy.}
After each terminal label, the reset policy decides whether to retry or reset:
a failed attempt retries directly if the subtask's starting conditions still
hold, whereas a successful attempt that changed them requires a reset, which
the robot performs when feasible, for example by lowering and releasing a
grasped object, and a human performs otherwise.

\subsection{Success-Reweighted Retraining and Redeployment}
\label{sec:retraining}

Online updates provide a weak learning signal for a bottleneck whose base success is
low: most attempts fail, so rewarded transitions are rare in the replay
buffer. A second problem arises from how local practice is reset. To save reset time, attempts are often restaged to the subtask's starting
conditions rather than to the task's initial state, so consecutive attempts
see nearly the same scene configuration; online updates can then overfit to
that configuration and fail under the state variation that the preceding
nominal behavior produces at evaluation.

\methodshort{} therefore periodically retrains each residual policy on a
success-reweighted copy of its replay and redeploys it~\cite{mark2023offline}. The curated dataset
keeps every successful episode and a uniformly sampled fraction $\rho$ of the
failed ones, $\widetilde{\mathcal D}_k=\mathcal D_k^{+}\cup
\operatorname{Sample}_\rho(\mathcal D_k^{-})$, which raises the share of
rewarded experience across all restagings while retaining some failures as
negatives. Fresh actor and critic networks are trained on this fixed dataset,
and an operator selects a candidate checkpoint. The selected checkpoint and curated replay buffer initialize a new online run, where updates resume as new rollouts are collected. Retraining thus changes the policy used for subsequent data collection, rather than serving solely as a final policy extraction step.

\section{Real-World Experiments}
\label{sec:experiments}

Our experiments address three questions. \textbf{Q1.} Does targeted subtask RL
improve a pretrained VLA on complete long-horizon tasks? \textbf{Q2.} Under a
matched robot-rollout budget, how does \methodshort{} compare with RL
fine-tuning methods that train on the full task? \textbf{Q3.} Does
success-reweighted retraining matter?

\subsection{Tasks and Setup}
\label{sec:tasks}

We evaluate \methodshort{} on three long-horizon tasks across two real-robot platforms (Fig.~\ref{fig:tasks}).

\begin{itemize}
  \item \textbf{Earbud insertion (bimanual YAM).}
  The robot opens a charging case, holds it in the left gripper, inserts two
  earbuds with the right arm, and closes the case. The task is hard because it
  is a dependent chain: a poor holding pose propagates to both insertions; each
  slot's clearance is small relative to the positioning error of bimanual
  coordination; and the gripper partly occludes the slot, so a seated earbud
  and one resting on the case look alike until release, which complicates both
  control and outcome assessment. The base policy reaches the second insertion
  only after earlier successes, so full-task rollouts rarely practice it. The
  bottlenecks are case preparation (RL0), first insertion (RL1), and second
  insertion (RL2).
  \item \textbf{LEGO sorting (bimanual YAM).}
  The robot sorts ten bricks into three color bins within 150\,s. The SFT
  demonstrations come from the public ABC-130k dataset~\cite{abc2026}, whose
  bricks are larger than ours, so the nominal gripper often closes too little
  to retain a brick even from a well-placed approach. Since the grasp recurs
  ten times per episode, a modest per-grasp failure rate consumes the time
  budget. The bottleneck is the grasp by either arm, and the base policy
  supplies approach, transport, and placement.
  \item \textbf{Cable unplug-and-plug (single-arm Franka FR3).}
  The robot unplugs a cable from a source router and inserts it into a
  destination router. Insertion demands alignment within the port tolerance
  followed by a decisive contact motion, and small offsets catch the connector
  on the housing. Insertion is evaluated in the states produced by the
  preceding extraction. The bottleneck is alignment and insertion.
\end{itemize}

These tasks involve grasping, repositioning, object handovers, and alignment, and span 20--120\,s (approximately 600--3,600 control steps at 30\,Hz). Bottleneck subtasks typically last 3--15\,s (90--450 control steps).

\textbf{Base policies and contracts.}
Humans identify the bottlenecks
from real-robot evaluations of the base policies, provide a few subtask
demonstrations to delimit them, and specify each contract.
Reward and reset protocols depend on how hard the outcome is to recognize
and how hard the scene is to restore (Table~\ref{tab:human_involvement}).
For the LEGO and cable tasks, a lifted brick or a seated connector is visually
unambiguous, so the automatic success verifier supplies every reward. For
earbuds, a seated earbud and one resting on the case look alike to a VLM
without task-specific post-training, so a human checks the verifier's labels.
Resets are automated when the robot can restore the initial conditions, as in the LEGO task, where it lowers and releases a grasped brick onto the table. Human assistance is required when a reset exceeds the robot's hardware capabilities, such as extracting a seated earbud from its case, or when the scene cannot be restored autonomously, such as after an earbud falls to the floor.

\textbf{Evaluation.}
We evaluate each method on 20 complete episodes per task, starting from the full-task's initial states.
To capture partial completion and improvements that binary full-task success can obscure in long-horizon tasks, we additionally report a normalized progress score. LEGO progress is the mean fraction of the ten bricks sorted before the deadline; because the task repeats one pick-and-place ten times, we report only this score for LEGO, as a binary outcome would reduce to whether a single brick is placed. Earbud progress assigns one third per completed stage, with success requiring all three stages. Cable progress assigns one half each for unplugging and plugging, with success requiring both.
We also report per-stage success within these same episodes as the fraction in which each bottleneck's outcome criterion is satisfied, independently of other stages' outcomes. For example, the second earbud may be seated correctly even if the first is misplaced inside the case.

\begin{figure*}[t]
  \centering
  \includegraphics[width=1\textwidth]{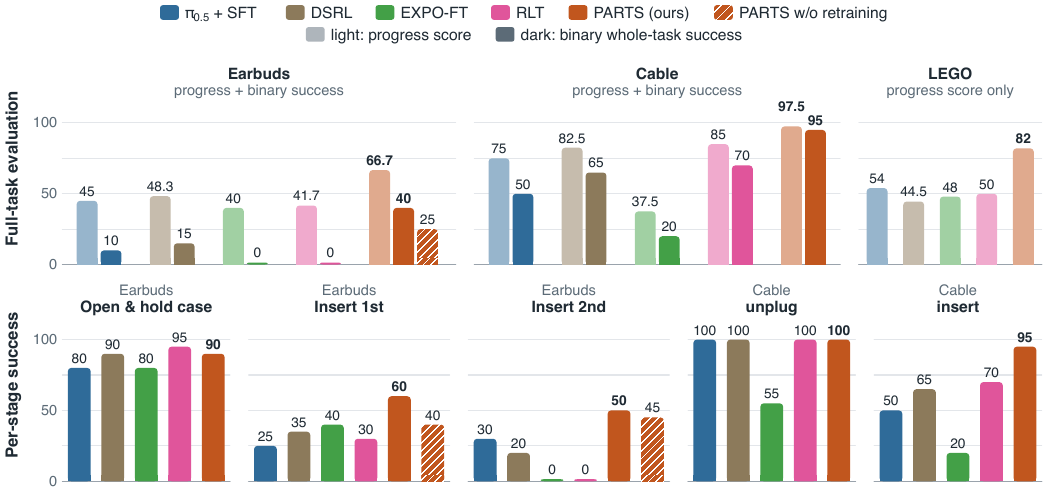}
  \caption{\methodshort{} significantly raises success across tasks with different difficulty levels and outperforms RL
  fine-tuning baselines under a matched robot-rollout budget. \emph{Top:}
  full-task evaluation over 20 episodes per method; for each method the light
  bar is the progress score and the dark bar is binary whole-task success.
  Earbud progress credits one third per completed stage and success requires
  all three; cable progress credits one half each for unplugging and insertion
  and success requires both. LEGO is a repeated pick-and-place task over ten
  bricks, so we report only its progress score, the fraction of bricks sorted
  before the deadline, which is more informative than a binary outcome over
  the whole task. The ablation is reported for binary success only.
  \emph{Bottom:} per-stage success measured within the same episodes, for the
  three earbud bottlenecks and the two cable stages. Hatched bars are \methodshort{}
  without success-reweighted retraining.}
  \label{fig:results}
  \vspace{-9pt}
\end{figure*}

\subsection{Baselines}
\label{sec:baselines}

We compare \methodshort{} to the frozen base policy and to three RL methods that
improve a pretrained VLA from real-world experience. For fair comparison, every
RL method starts from the same $\pi_{0.5}$-SFT base policy, trains with the
same amount of robot rollout time, and receives no corrective
teleoperation or DAgger-style~\cite{kelly2019hg} human interventions during RL rollouts. All baselines use human resets and human reward labeling; on LEGO the label is the progress score, the fraction of bricks sorted in the episode.
\begin{itemize}
  \item \textbf{SFT}: We report the performance of the base policy after SFT on task demonstrations, before any online interaction.
  \item \textbf{DSRL}~\cite{wagenmaker2025steering}: DSRL learns an online RL policy
  in the latent noise space of the frozen VLA, steering action generation
  by selecting the noise fed to the VLA's action generator. Exploration is
  thereby confined to actions the VLA can generate. We run DSRL on the full
  task.
  \item \textbf{EXPO-FT}~\cite{dong2026expo}: EXPO-FT learns an edit policy that
  modifies the VLA's action chunks under Q-guidance while continuing to update
  the $\pi_{0.5}$ backbone. Unlike \methodshort{}, it adapts the base policy
  itself and operates on the full task. Its published regime additionally uses
  human interventions and automatic reward detectors, which we do not use.
  \item \textbf{RLT}~\cite{xu2026rl}: like \methodshort{}, RLT freezes the VLA and
  trains a small actor-critic that refines the VLA's reference action chunks,
  conditioned on the VLA's representation. Unlike \methodshort{}, RLT has a
  single RL phase entered by a human-selected VLA-to-RL handoff, after which
  the RL policy controls the rest of the episode with no handback to the VLA.
  We follow this procedure on earbuds and cable. It cannot express LEGO
  sorting, whose grasp bottleneck recurs for every brick and requires repeated
  switching between the VLA and the residual, so on LEGO we run RLT as
  full-task RL. RLT's optional corrections are not used.
\end{itemize}

\textbf{Matched rollout budget.}
Each baseline receives the same RL robot-rollout time as \methodshort{} on the
same task, counted as elapsed robot time during RL rollouts, excluding
physical resets and pauses for RL updates. DSRL and EXPO-FT spend
this budget on full-task rollouts, RLT spends it from its handoff onward on earbuds and
cable and on the full task on LEGO, and \methodshort{} spends it inside the
bottlenecks. The comparison therefore controls robot data-collection time,
while wall-clock and compute costs can differ.

\textbf{Implementation details.}
We use official DSRL and EXPO-FT code on Franka and reimplement both for YAM, changing only the robot interfaces. RLT is reproduced from the paper, as no official code is available.

\subsection{Experimental Results}
\label{sec:results}

\textbf{Q1: \methodshort{} improves over the base VLA policy.}
Fig.~\ref{fig:results} shows that \methodshort{} improves performance in full-task evaluations across all three long-horizon tasks and both robot platforms, including settings with weak base policies. On the relatively easy cable task, insertion success nearly doubles after only 29\,min of RL rollouts.
This training efficiency stems from our system design: because the base policy already unplugs reliably, RL is concentrated on insertion, directing online interaction toward the subtask that limits success.
On the most challenging earbud task, which requires precise bimanual manipulation, \methodshort{} achieves four times the base policy's full-task success rate. Per-stage results show substantial gains in both insertions, whose success rates improve by 20-35 percentage points over the base policy.
For LEGO, the longest task at 120\,s, correcting gripper closure alone increases progress from 54\% to 82\% with only 17\,min of RL rollouts.
The same two grasp residual policies are reused across all ten bricks, allowing each learned correction to address repeated occurrences of its bottleneck throughout the task.

\textbf{Q2: \methodshort{} outperforms RL fine-tuning baselines under the same
rollout budget.}
Given the same robot rollout time, \methodshort{} consistently outperforms all baselines across tasks of varying difficulty. It concentrates learning and repeated practice on bottleneck subtasks, whereas the baselines spend part of their rollout budget re-executing subtasks that the base policy already handles reliably.
Moreover, the baselines use sparse terminal rewards that provide outcome feedback only at the end of each 20--120\,s episode. Low base-policy success rates leave the replay buffer with few successful trajectories, making critic learning difficult under sparse positive feedback and long credit-assignment horizons.
In contrast, \methodshort{} uses local rewards over 3--15\,s subtask windows with initial success rates above 25\%, providing more frequent positive feedback and shortening the credit-assignment horizon.
EXPO-FT ultimately underperforms the base policy, occasionally degrading previously reliable behaviors such as cable unplugging.
This suggests that backbone updates driven by sparse full-task rewards can disrupt reliable behaviors in non-bottleneck subtasks.
RLT and DSRL are stronger baselines, but neither significantly improves on the SFT policy on average under limited real-world rollout budgets. DSRL constrains exploration to behaviors generated by the frozen VLA, promoting stable learning but potentially limiting further improvements.
RLT focuses on critical-phase adaptation, but still underperforms \methodshort{}. Its reported system targets a single critical phase following a VLA-to-RL handoff, whereas \methodshort{} coordinates repeated transitions between the base policy and specialized residual policies to address multiple bottlenecks in the LEGO and earbud tasks.

\textbf{Q3: success-reweighted retraining matters.}
We ablate retraining on the earbud task, where it was applied to RL1 and RL2
only. Both start from low base success, so their local
rewards are sparse, and both are restaged to RL0's success state between
attempts. Restaging removes the variation that full-task execution introduces
into the relative pose of the held case and the inserting gripper, so
consecutive attempts see similar configurations, and a residual updated
online alone can settle into corrections that fit that configuration but not
the states produced by the preceding nominal behavior at evaluation. RL0 was
not retrained: its base success is already high and its reset is the
task's initial condition, so its attempts already cover the evaluation
distribution. Removing success-reweighted retraining reduces both subtask and full-task success, as shown in the hatched bars of Fig.~\ref{fig:results}.
Therefore, retraining fresh networks on the curated episode set can help escape local optima,
providing a key mechanism underlying \methodshort{}'s success.

\section{CONCLUSION}
\label{sec:conclusion}

We presented \methodshort{}, a real-world subtask RL framework that adapts a
pretrained robot policy to a long-horizon task by concentrating practice on the
few subtasks where the policy fails, with minimal human intervention during
training. Across bimanual and single-arm platforms, \methodshort{} substantially improves full-task success over the base policy and outperforms existing RL fine-tuning methods under matched robot rollout budgets, while requiring fewer recurring forms of human
involvement.

Future work will pursue full autonomy through agents that identify bottlenecks from policy failures, generate dense subtask rewards, and construct recovery phases. These capabilities could automate setup decisions, accelerate learning when successful outcomes are rare, and enable the robot to practice recovery behaviors after subtask failures, reducing reliance on human resets.

\bibliographystyle{IEEEtran}
\bibliography{references}

\clearpage
\onecolumn
\appendix

\subsection{Robot Platforms}
\label{app:platforms}

\begin{figure}[h]
  \centering
  \begin{minipage}[t]{0.3\linewidth}
    \centering
    \includegraphics[width=\linewidth]{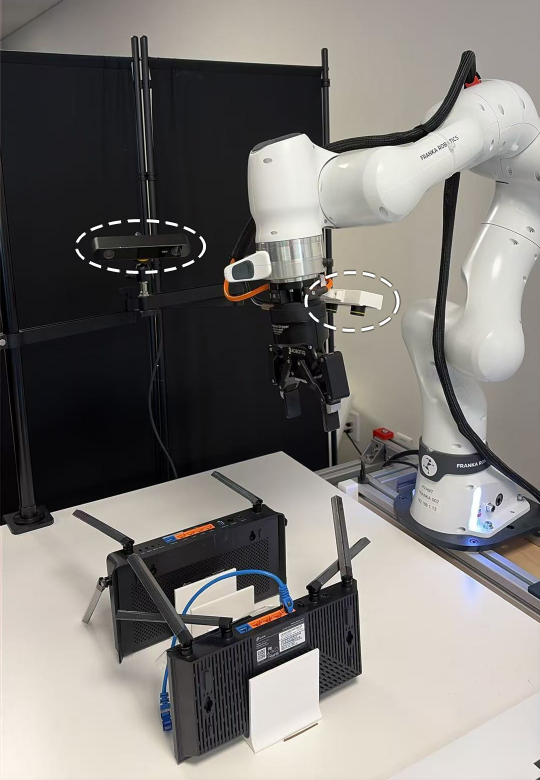}\\[3pt]
    {\footnotesize (a) Single-arm Franka FR3}
  \end{minipage}\hspace{0.05\linewidth}%
  \begin{minipage}[t]{0.3\linewidth}
    \centering
    \includegraphics[width=\linewidth]{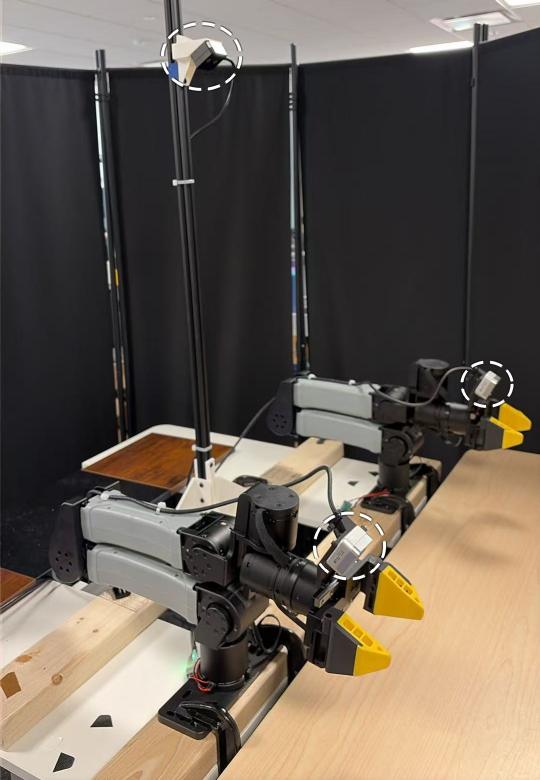}\\[3pt]
    {\footnotesize (b) Bimanual YAM}
  \end{minipage}
  \caption{Robot platforms. (a) Franka FR3, shown with the two routers of the
  cable task, observed by a fixed ZED camera and a wrist-mounted ZED camera.
  (b) YAM with I2RT FlexPoint adaptive grippers and three Intel RealSense D405
  cameras: a head camera on the overhead pole and one camera on each wrist. Dashed ellipses mark the cameras.}
  \label{fig:platforms}
\end{figure}

Fig.~\ref{fig:platforms} shows the two robot platforms. The single-arm Franka
FR3 carries a Robotiq gripper and is observed by two ZED stereo cameras, one on
a fixed stand and one mounted at the wrist; it runs the cable unplug-and-plug
task. The bimanual YAM station has two 6-DoF YAM arms with I2RT FlexPoint
adaptive grippers, whose compliant fingertips conform to the grasped object.
Three Intel RealSense D405 cameras observe it: a head camera on an overhead pole
and one camera on each wrist. It runs earbud insertion and LEGO sorting, with the same
one-head, two-wrist camera layout as the ABC-130k stations
(Appendix~\ref{app:sft}).

\subsection{SFT Datasets}
\label{app:sft}

Each base policy $\pi_0$ is $\pi_{0.5}$~\cite{pmlr-v305-black25a} after supervised fine-tuning on a
task-specific demonstration set $\mathcal{D}_{\text{exp}}$
(Sec.~\ref{sec:problem}), and every RL method in Sec.~\ref{sec:experiments}
starts from it. Table~\ref{tab:sft_data} lists these sets. Both YAM tasks draw
on ABC-130k~\cite{abc2026}, a public teleoperation dataset recorded on bimanual
YAM stations. For earbud insertion we add our own teleoperated demonstrations,
and the cable task uses only our own.

\begin{table}[h]
  \centering
  \caption{SFT Demonstrations for Each Base Policy}
  \label{tab:sft_data}
  \renewcommand{\arraystretch}{1.3}
  \begin{tabular}{@{}p{0.17\linewidth}p{0.19\linewidth}p{0.10\linewidth}p{0.45\linewidth}@{}}
    \toprule
    \textbf{Task} & \textbf{Resource} & \textbf{Episodes} & \textbf{Link} \\
    \midrule
    Earbud insertion\newline(bimanual YAM)
      & ABC-130k episodes & 244 of 2,095
      & \url{https://huggingface.co/datasets/XDOF/ABC-130k/tree/main/data/train/insert_the_wireless_bluetooth_earbuds_into_the_charging_case} \\
      & Our teleoperation & 38
      & \url{https://huggingface.co/datasets/Sichang0621/earbuds_30fps} \\
      & Merged training set & 282
      & \url{https://huggingface.co/datasets/Sichang0621/earbuds_teleop_abcyellow_v21} \\
    \midrule
    LEGO sorting\newline(bimanual YAM)
      & ABC-130k episodes & 500 of 4,458
      & \url{https://huggingface.co/datasets/XDOF/ABC-130k/tree/main/data/train/sort_the_legos_into_containers_by_color} \\
      & Converted training set & 500
      & \url{https://huggingface.co/datasets/Sichang0621/abc_sort_legos_v21} \\
    \midrule
    Cable unplug-and-plug\newline(Franka FR3)
      & Our teleoperation & 100 & -- \\
    \bottomrule
  \end{tabular}
\end{table}

\noindent\textbf{Episode selection.}
For LEGO sorting, we take the first 500 of the task's 4,458 ABC-130k training
episodes in order of episode UUID, which amounts to a random subset. For earbud
insertion, we take 244 of the task's 2,095 training episodes and merge them with
38 demonstrations teleoperated on our YAM setup; the merged set of 282 episodes
is the earbud training set.

\medskip
\noindent\textbf{Format.}
ABC-130k (\url{https://huggingface.co/datasets/XDOF/ABC-130k}) stores each
episode as MCAP logs whose streams are recorded on independent clocks. We
convert the selected episodes to the LeRobot dataset format (v2.1) for
fine-tuning, resampling all streams onto a common 30\,Hz grid.
Both YAM training sets use this format, with 14-D states and actions (six joint
positions and one gripper value per arm) and three $640\times480$ RGB streams
from a head camera and two wrist cameras. The LEGO training set contains
10.7\,h of robot data and the earbud training set 8.5\,h, of which our
teleoperation contributes 32\,min.

\subsection{Additional Experiment Results}
\label{app:results}

Table~\ref{tab:cable_time} complements the cable results in
Fig.~\ref{fig:results} with the mean full-task completion time, measured on the
same 20 evaluation episodes per method (Sec.~\ref{sec:tasks}). Completion time
is averaged over each method's successful episodes only, so the means rest on
between 4 (EXPO-FT) and 19 (\methodshort{}) episodes. \methodshort{} attains
both the highest success rate and the lowest mean completion time, 20.1\,s
compared with 35.8\,s for the SFT base policy. DSRL and RLT also complete the
task faster than SFT on average, while EXPO-FT is both slower and less
successful.

\begin{table}[h]
  \centering
  \caption{Full-Task Success and Completion Time on Franka Cable Unplug-and-Plug}
  \label{tab:cable_time}
  \begin{tabular}{lcc}
    \toprule
    Method & Success rate $\uparrow$ & Mean completion time (s) $\downarrow$ \\
    \midrule
    SFT & 50\% (10/20) & 35.8 \\
    DSRL~\cite{wagenmaker2025steering} & 65\% (13/20) & 21.0 \\
    EXPO-FT~\cite{dong2026expo} & 20\% (4/20) & 41.1 \\
    RLT~\cite{xu2026rl} & 70\% (14/20) & 24.0 \\
    \midrule
    \methodshort{} (ours) & \textbf{95\%} (19/20) & \textbf{20.1} \\
    \bottomrule
  \end{tabular}
\end{table}

\clearpage
\subsection{Human Contracts and Executable Scaffolding}
\label{app:scaffolding}

Repeated local practice requires an explicit definition of where an attempt
starts, what RL may change, what counts as success, and how to prepare the next
attempt. This section illustrates how a human contract supplies these choices
and how the coding-agent workflow turns them into a policy selector, a success
verifier, and an auto-reset policy. We use the right-arm LEGO grasp as a concrete
example; the frozen VLA supplies approach, lifting, transport, and placement,
while the residual corrects gripper openness.

\medskip
\noindent\textbf{Human Contract: Right-Arm LEGO Grasp.}
The following is a human-readable summary of the gripper-only configuration,
with automatic verification and in-place training resets. It states the desired
behavior rather than prescribing a grasp trajectory.

\smallskip
\begingroup
\renewcommand{\arraystretch}{1.3}
\noindent\begin{tabular}{@{}p{0.19\linewidth}p{0.77\linewidth}@{}}
\toprule
\textbf{Contract field} & \textbf{Specification} \\
\midrule
Entry states & The selected gripper is open and in the tabletop grasp band;
the right wrist view confirms a LEGO brick near the gripper center, outside
the sorting containers. Require consecutive qualifying observations before
activating the residual. \\
Correctable actions & Correct right-gripper openness only:
$M=(0,0,0,0,0,0,1)$ in the selected arm's seven action coordinates.
This configuration bounds the residual by $\pm0.50$ and clips final
normalized openness to $[0,1]$; the other coordinates follow the frozen VLA. \\
Success criterion & Lift the brick at least $0.05$\,m above the attempt's
recorded grasp height and retain it for at least $1.0$\,s. Both the physical
lift and fresh wrist-image evidence of holding must persist. \\
Reset goal & During local training, lower a held brick at its current planar
position to the recorded grasp height, then release and reopen. After an empty
or missed grasp, reopen and retry if the entry conditions remain valid. \\
Termination & End an attempt on its terminal outcome or configured motion
budget. Reset execution also has a finite budget and exposes failure to the
runtime if it cannot complete. \\
Full-task evaluation & After verified grasp success, return to the VLA for
transport and placement; do not perform the training place-back reset. \\
\bottomrule
\end{tabular}
\endgroup

\medskip
Humans provide the contract, example demonstrations, and labeled episodes for
review. Coding agents implement and revise the three programs against the robot
interfaces and this feedback. The deployed programs then run as ordinary code;
no coding-agent call is required at each control step. The LEGO implementation
uses SAM3-based visual evidence. VLM-based checks are supported elsewhere in
the system, but are not required by this example.

\definecolor{codebg}{HTML}{D9D9D9}
\definecolor{codekeyword}{HTML}{0000FF}
\definecolor{codecontrol}{HTML}{AF00DB}
\definecolor{codefunction}{HTML}{795E26}
\definecolor{codetype}{HTML}{267F99}
\definecolor{codevariable}{HTML}{001080}
\definecolor{codestring}{HTML}{A31515}
\definecolor{codenumber}{HTML}{098658}
\definecolor{codecomment}{HTML}{008000}
\lstdefinestyle{partsappendix}{
  language=Python,
  basicstyle=\fontencoding{T1}\fontfamily{fvm}\selectfont\def\ttdefault{fvm}\small,
  upquote=true,
  identifierstyle=\color{codevariable},
  keywords=[1]{def,class,lambda,and,or,not,in,is,None,True,False,self,global,nonlocal},
  keywordstyle=[1]\color{codekeyword},
  keywords=[2]{if,elif,else,for,while,break,continue,return,pass,raise,try,except,%
    finally,with,as,import,from,yield,assert,del,async,await},
  keywordstyle=[2]\color{codecontrol},
  keywords=[3]{abs,all,any,enumerate,getattr,hasattr,isinstance,len,max,min,print,%
    range,round,sorted,sum,zip,array,update,hold,_open,_gripper_above_lego,%
    move_right_ee_to},
  keywordstyle=[3]\color{codefunction},
  keywords=[4]{bool,float,int,str,list,dict,tuple,set,np},
  keywordstyle=[4]\color{codetype},
  commentstyle=\color{codecomment},
  stringstyle=\color{codestring},
  deletecomment=[s]{"""}{"""},
  deletecomment=[s]{'''}{'''},
  morestring=[s]{"""}{"""},
  morestring=[s]{'''}{'''},
  literate=*{0}{{\color{codenumber}0}}1 {1}{{\color{codenumber}1}}1
    {2}{{\color{codenumber}2}}1 {3}{{\color{codenumber}3}}1
    {4}{{\color{codenumber}4}}1 {5}{{\color{codenumber}5}}1
    {6}{{\color{codenumber}6}}1 {7}{{\color{codenumber}7}}1
    {8}{{\color{codenumber}8}}1 {9}{{\color{codenumber}9}}1
    {.0}{{\color{codenumber}.0}}2 {.1}{{\color{codenumber}.1}}2
    {.2}{{\color{codenumber}.2}}2 {.3}{{\color{codenumber}.3}}2
    {.4}{{\color{codenumber}.4}}2 {.5}{{\color{codenumber}.5}}2
    {.6}{{\color{codenumber}.6}}2 {.7}{{\color{codenumber}.7}}2
    {.8}{{\color{codenumber}.8}}2 {.9}{{\color{codenumber}.9}}2,
  showstringspaces=false,
  columns=fullflexible,
  keepspaces=true,
  breaklines=true,
  aboveskip=0pt,
  belowskip=0pt
}
\newtcblisting{partscode}{listing only, listing options={style=partsappendix},
  colback=codebg, colframe=codebg, boxrule=0pt, arc=0pt, left=6pt, right=6pt,
  top=4pt, bottom=4pt, before skip=7pt, after skip=7pt}

\medskip
\noindent\textbf{Policy Selector: Enter the Bottleneck.}
The selector requests RL only after the entry predicate holds repeatedly. The
excerpt below is inside the VLA-mode branch; \texttt{None} leaves the mode
unchanged. The omitted helper checks robot state, grasp height, gripper opening,
wrist-image centering, and exclusion of sorting containers. The runtime uses
the success verifier, rather than this entry counter, as the authoritative
success signal.

\begin{partscode}
if self._gripper_above_lego(obs):
    self._above_count += 1
else:
    self._above_count = 0
if self._above_count >= self.start_frames:
    return "rl"
return None
\end{partscode}

\newpage
\noindent\textbf{Success Verifier: Require a Lift and Fresh Holding Evidence.}
Before the following fragment, the verifier checks that the current grasp
attempt has reached the required lift height and rejects missing, stale, pre-lift, or out-of-order
image evidence. It then measures the span of positive \emph{capture timestamps},
not the number of times a cached prediction is read. A negative observation
clears the positive interval. Success requires both this interval and the
continuous physical lift to last for \texttt{stable\_hold\_s} (1.0\,s in the
illustrated configuration).

\begin{partscode}
held = bool(held_raw["held"])
result.update(grasp_held=held, grasp_held_sample_time=capture)
if not held:
    self._positive_capture_first = self._positive_capture_last = None
    return result
if fresh:
    last = self._positive_capture_last
    if last is None or capture - last > self.held_sample_max_age_s:
        self._positive_capture_first = capture
    self._positive_capture_last = capture
first, last = self._positive_capture_first, self._positive_capture_last
if first is not None and last is not None:
    elapsed = last - first
    result.update(
        grasp_stable_elapsed_s=elapsed,
        success=(elapsed >= self.stable_hold_s
                 and now - self._lift_stable_since >= self.stable_hold_s))
return result
\end{partscode}

\medskip
\noindent\textbf{Auto-Reset Policy: Lower, Release, and Repeat.}
For the in-place training reset, the program preserves the grasp while lowering
the brick, then opens the gripper after reaching the target height or detecting
contact near it. The fragment begins inside the lowering branch.
Unlike the selector and verifier, which return decisions and evidence, the
reset program invokes existing motion primitives to move the robot.

\begin{partscode}
    reached = z <= target_z + 0.008
    if not (reached or contact):
        step_down = 0.010 if z <= target_z + 0.04 else 0.012
        move_right_ee_to(
            (tx, ty, float(max(float(target_z), z - step_down))),
            max_steps=60)
        return hold()
    self.phase = "open"

if self.phase == "open":
    self._open()
    self.release_steps += 1
\end{partscode}

\medskip
\noindent\textbf{Scope of the excerpts.}
These fragments omit initialization, calibration, helper functions, and runtime
budget/failure handling; they are not standalone controllers. Indentation and
line breaks are adjusted, and reset comments are omitted; executable logic is
unchanged. Values are implementation settings, not additional experimental
measurements. The source also supports a human-label development mode, in which
automatic evidence does not override the human reward label. Reported task
protocols remain those in Sec.~\ref{sec:tasks}: automatic rewards for LEGO and
cable, human checks for earbud rewards, and automatic physical resets for LEGO.
None of these protocols uses corrective human actions during RL rollouts.

\end{document}